\documentclass[10pt,english]{article}

\usepackage{geometry}
\usepackage[T1]{fontenc}
\usepackage[justification=centering]{caption}

\usepackage{amsmath,amssymb}
\usepackage{multirow}

\usepackage{changepage}

\usepackage[utf8x]{inputenc}

\usepackage{textcomp,marvosym}

\usepackage{cite}

\usepackage{nameref,hyperref}

\usepackage[table]{xcolor}

\usepackage{array}

\newcolumntype{+}{!{\vrule width 2pt}}

\newlength\savedwidth

\usepackage[aboveskip=1pt,labelfont=bf,labelsep=period,justification=raggedright,singlelinecheck=off]{caption}
\renewcommand{\figurename}{Fig}

\makeatletter
\renewcommand{\@biblabel}[1]{\quad#1.}
\makeatother

\usepackage{epstopdf}

\usepackage[pdftex]{graphicx}

\begin{document}
\vspace*{0.2in}

\begin{flushleft}
{\Large
\textbf\newline{Image-based Plant Disease Diagnosis with Unsupervised Anomaly Detection Based on Reconstructability of Colors} 
}
\newline
\\
Ryoya Katafuchi \textsuperscript{1\Yinyang},
*Terumasa Tokunaga\textsuperscript{1\Yinyang},
\\
\bigskip
\textbf{1} Department of Systems Design and Informatics, Kyushu Institute of Technology, Iizuka, Fukuoka, JAPAN
\\
\bigskip

%
%
\Yinyang 




* tokunaga@ces.kyutech.ac.jp

\end{flushleft}
\section*{Abstract}
This paper proposes an unsupervised anomaly detection technique for image-based plant disease diagnosis.
The construction of large and publicly available datasets containing labeled images of \textit{healthy} and \textit{diseased} crop plants led to growing interest in computer vision techniques for automatic plant disease diagnosis. Although supervised image classifiers based on deep learning can be a powerful tool for plant disease diagnosis, they require a huge amount of labeled data.
The data mining technique of anomaly detection includes unsupervised approaches that do not require rare samples for training classifiers.
We propose an unsupervised anomaly detection technique for image-based plant disease diagnosis that is based on the reconstructability of colors; a deep encoder-decoder network trained to reconstruct the colors of \textit{healthy} plant images should fail to reconstruct colors of symptomatic regions.
Our proposed method includes a new image-based framework for plant disease detection that utilizes a conditional adversarial network called pix2pix and a new anomaly score based on CIEDE2000 color difference.
Experiments with PlantVillage dataset demonstrated the superiority of our proposed method compared to an existing anomaly detector at identifying diseased crop images in terms of accuracy, interpretability and computational efficiency.


\section*{\uppercase{Introduction}}
Plant disease diagnosis is an important task for food safety and security.
The PlantVillage project~\cite{Hughes_2016} was started to develop accurate image classifiers for plant disease diagnosis. This is a publicly available image dataset for developing automatic diagnostic techniques to identify plant diseases. It provides thousands of labeled images of healthy and diseased crop plants collected under controlled conditions. Such a large dataset has been used to establish deep learning challenges for developing an accurate image classifier for plant disease diagnosis.

In a comprehensive experiment using color crop images, AlexNet and GoogleNet achieved average accuracies of over 90\% at identifying 26 diseases in 14 crop species~\cite{Mohanty_2016}. Similarly, LeNet accurately classified diseased banana leaves under severe conditions~\cite{Amara2017}. Ferentinos~(2018)~\cite{Ferentinos_2018} compared the performances of five convolutional neural network (CNN) models using leaf images obtained under both laboratory and real conditions. The VGG model achieved the best performance with a success rate of 99.53\%. Another study using the PlantVillege dataset showed that deep learning provided outstanding performance compared with conventional machine learning techniques~\cite{Radovanovic_2020}.

The results of the previous studies warrant further developments of image-based plant disease diagnosis techniques for more practical applications. Supervised image classifiers based on deep learning require a huge amount of labeled data for training. Correcting samples with rare diseases often imposes a severe burden on human annotators, which can be a severe bottleneck for practical application. To address this issue, further studies are needed to develop an image-based diagnosis technique that is free of annotation costs for rare samples.

Anomaly detection is a data mining technique for identifying irregular or unusual patterns in datasets. This technique exhibits a wide range of applications, such as fraud detection for financial services, intrusion detection for networks, identification of disease markers for medical diagnosis, and failure detection for engineering systems. Typical approaches to anomaly detection are based on conventional machine learning. Simple clustering approaches are often used for unlabeled data ~\cite{Xiong_2011} \cite{Zimek_2012}. In cases where normal and anomalous labels are available, simple classification approaches such as support vector machines are used~\cite{Chen_2001}.

Recently, many anomaly detection techniques based on deep neural networks have been proposed in the fields of machine learning and computer vision. Deep anomaly detection can be categorized into three groups based on the type of machine learning: supervised approaches~\cite{Chalapathy_2017}, unsupervised approaches~\cite{Patterson_2017}, {\cite{Tuor_2017}, \cite{Sutskever_2008}, {\cite{Vincent_2008}, {\cite{Rodriguez_1999}, {\cite{Lample_2016}}, and semi-supervised approaches~\cite{Edmunds_2017}, \cite{Racah_2017}, \cite{Perera_2019}. \cite{Chalapathy_2019} provides a comprehensive review of these approaches.

Generative adversarial networks (GANs)~\cite{Goodfellow_2014} demonstrated a great deal of success with image generation tasks~\cite{Radford_2016} \cite{Chen_2016} \cite{Salimans_2016} \cite{Gulrajan_2017} \cite{Mao_2017} \cite{Isola2017}. The excellent expressive power of GANs led to growing interest in utilizing them for real-world data analysis, including anomaly detection. GANs are typically applied to adversarial feature learning~\cite{Donahue_2016} of normal data and measuring anomaly scores for a given query image. AnoGAN~\cite{Schlegl2017} provides an unsupervised approache to detecting real-world anomalies, including the discovery of novel anomalies in medical imaging data.
Recently, some extensions of AnoGAN have been proposed to overcome performance issues~\cite{Zenati_2018} \cite{Akcay_2018} or improve the computational efficiency~\cite{Schlegl_2019}.

AnoGAN computes the anomaly score based on the reconstructability of normal samples. Because AnoGAN does not require anomalous data for training neural networks, it is applicable to diverse problems, including those within the natural sciences. However, this approach does not explicitly focus on colors in imaging data. In many real-world problems, color information is essential to discovering anomalies in datasets. For example, discoloration of leaves can be crucial information for detecting symptoms ~\cite{Riley_2002}.

To the best of our knowledge, there have been no extensions of AnoGAN focused on detecting color anomalies. Moreover, AnoGAN exhibits two drawbacks for real-time applications: it requires a huge amount of normal data to learn a manifold of normal variability, and it requires an iterative procedure for calculating anomaly scores, which reduces computational efficiency.

In this paper, We propose a new anomaly detection method for detecting plant diseases at the image level and visualizing symptomatic regions at the pixel level. The proposed method uses a conditional adversarial network called pix2pix~\cite{Isola2017} for learning inverse mapping from converted grayscale images to original color images. The simplicity of this strategy means that the proposed method should work well even in cases where a large amount of normal data is unavailable, unlike AnoGAN. We applied the proposed method to the PlantVillage dataset to explore its utility. Also, we propose a simple, and easy-to-interpret anomaly score that is based on the CIEDE2000 color difference. Because the proposed method does not require any iterative procedure for calculating the anomaly score, the computational efficiency is expected to be sufficient for real-time disease detection.

\section*{\uppercase{Related Work}}
The present work was motivated by AnoGAN~\cite{Schlegl2017} and its extensions~\cite{Zenati_2018}~\cite{Akcay_2018}~\cite{Schlegl_2019}. AnoGAN relies on the concept of reconstructing normal data from latent variables. This framework is applicable to diverse problems, but its effectiveness at detecting color anomalies has not been demonstrated in  previous studies. Our proposed method relies on the reconstruct colors. Specifically, we hypothesized that a rich generative model trained to reconstruct colors of normal data will fail to color anomalous regions in images. Unlike AnoGAN and its extensions, our focused was on detecting color anomalies, such as discolored parts on plants. Thus, the proposed method can be viewed as an extension of AnoGAN but in a different direction from previous studies.

\section*{\uppercase{Method}}
\subsection*{Outline of the Proposed Method}
Fig.~\ref{fig:Model} shows a schematic of the proposed method.
It consists of five steps for image-level detection and pixel-level visualization of plant diseases:
\begin{enumerate}
    \item  \textbf{Preparation: }
    Anomalies in color images are detected in terms of the reconstructability of colors at the pixel level with a conditional adversarial network.
    Consider a set of $M$ pairs of color images and grayscale images.
    Let $\mathbf{I}_c(x_i) \in \mathbb{R}^3_+ ~(i = 1, 2, \ldots, 256 \times 256)$ be a pixel value representing a color at the pixel position $x_i$.
    Similarly, let $I_g (x_i) \in \mathbb{R}_+ ~(i = 1, 2, \ldots, 256 \times 256)$ be a pixel value representing the intensity at the pixel position $x_i$.
    For notational simplicity, color and grayscale images are expressed as $\mathbf{I}_c$ and $\mathbf{I}_g$, respectively.
    Pairs of images, $\mathbf{I}_c$ and $\mathbf{I}_g$ are used to train the GAN so that the DCED network learns the inverse mapping $G: \mathbf{I}_g \mapsto \mathbf{I}_c$.
    These training images are selected only from normal data, while the test data include both normal and anomalous data.
    For evaluation purposes, we use a set of $N$ color images with an array of binary image-wise ground-truth labels $l_{n} \in \{1, -1\}~(n=1,2,\ldots,N)$.
    \\
    \item  \textbf{Training: }
   The left half of Fig.~\ref{fig:Model} illustrates concurrent training on generator and discriminator networks.
   Pairs of images $  \mathbf{I}_{g}$ and $\mathbf{I}_{c}$ are used as the input and output respectively of G.
   Pairs of $\mathbf{I}_{c}$ and $G(\mathbf{I}_g)$ are used as \textit{fake} pairs to train the discriminator network $D$.
       \\
    \item \textbf{Color reconstruction: }
    For the test data, the reconstructed color image $\mathbf{\hat{I'}}_c$ is obtained from a query color image $\mathbf{I'}_c$ by using the trained generator network.
    \\
    \item \textbf{Calculation of color anomaly score: }
    The anomaly score is calculated for a given query color image $\mathbf{\tilde{I}}_c$ based on color differences between the reconstructed color image $G(\mathbf{I'}_c)$ and the original color image $\mathbf{I'}_c$.
    The color difference $d_i$ is calculated from $\mathbf{I'}_c(x_i)$ and $\mathbf{\hat{I'}}_c(x_i)$ for all $i$.
    The anomaly score based on color difference is obtained simply by summing  $d_i$ for all $i$.
   \\
    \item \textbf{Anomaly detection: }
     Finally, a query image is classified as normal or anomalous by a simple thresholding of the anomaly score.
\end{enumerate}

\begin{figure}[htb]
\includegraphics[width=\hsize]{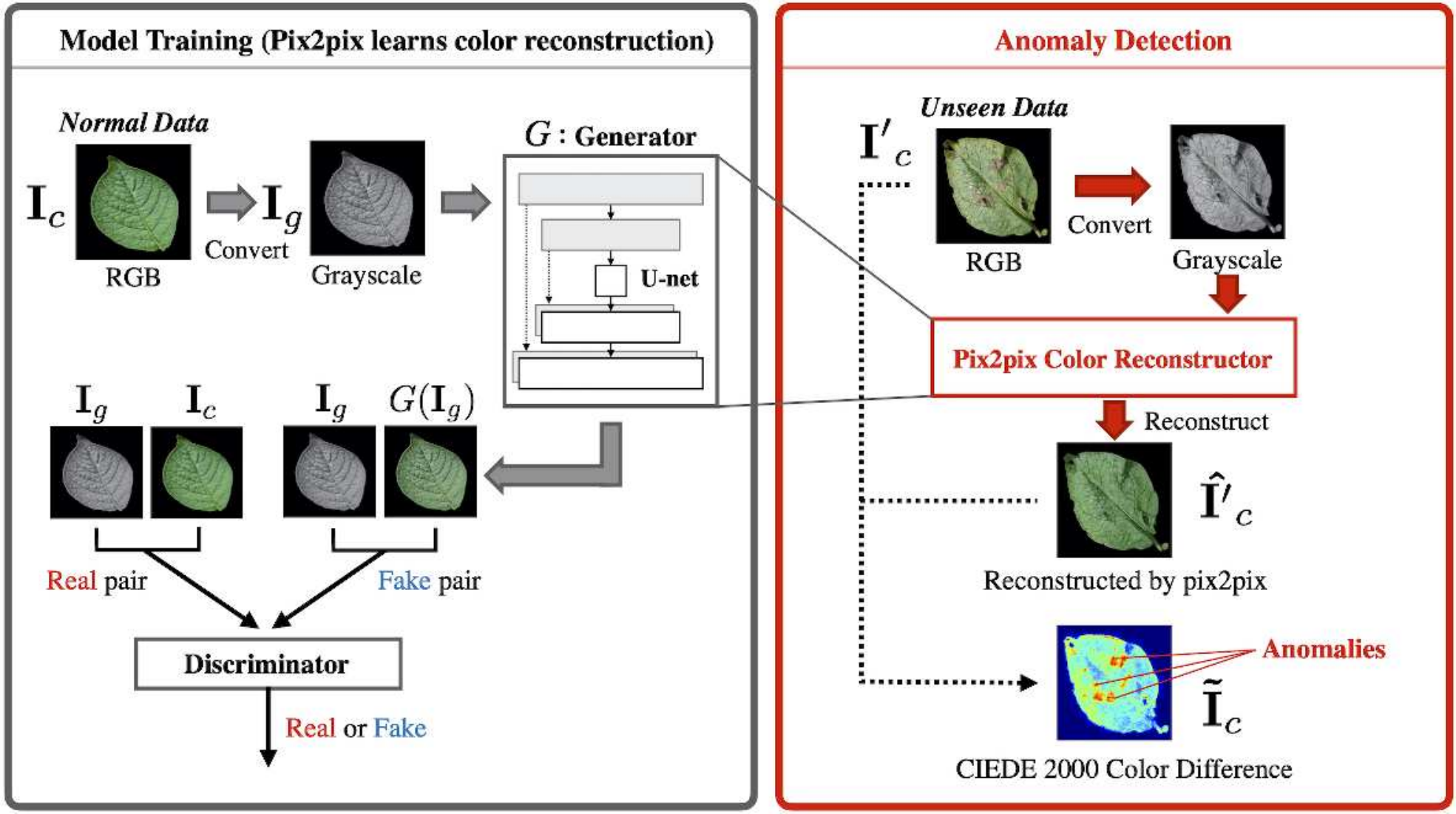}

\caption{Outline of the proposed method.}
\label{fig:Model}
\end{figure}

\subsection*{Color Reconstruction by Pix2pix}
We used pix2pix~\cite{Isola2017} for color reconstruction. Pix2pix is a general framework for image-to-image translation based on a deep convolutional GANs (DCGANs). The generator network is U-Net, which is a DCED network with skip structures. Skip structures enable a DCED network to learn both global and local features efficiently. The discriminator is the convolutional PatchGAN, which only penalizes the structure at the patch scale.

Let the input and output variables for pix2pix be $x$ and $y$, respectively.  Now, the loss function for training pix2pix can be expressed as follows:
\begin{eqnarray}
	\label{eq:objective}
 	L_{cGAN}(G, D) =\mathbb{E}_{x,y}[\log{D(x,y)}] +  \mathbb{E}_{x,z}[\log{\left(1-D(x,G(x,z) \right)}]
\end{eqnarray}
where $\mathbb{E}[ \cdot ]$ indicates the expected value and $z$ is a random noise vector.  During training, pix2pix seeks to minimise $L_{cGAN}$ with respect to $G$, and at the same time maximise it with respect to $D$ under $L_{1}$-regularization:
\begin{eqnarray}
	\label{eq:optimal}
 	G* &=& \arg \min_{G} \max_{D} \mathbb{E}_{x,y}[\log{D(x,y)}] +  \mathbb{E}_{x,z}[\log{\left(1-D(x,G(x,z) \right)}] \nonumber \\
	&+& \lambda  \mathbb{E}_{x,y,z} [ \| y - G(x,z)\|_1]
\end{eqnarray}

\subsection*{CIEDE2000 Color Anomaly Score}
For a given query color image, the anomaly score is calculated for each pixel. We propose a new anomaly score based on CIEDE2000, which reflects differences as perceived by humans. This should align the provides anomaly score with visual inspection by human. The CIEDE2000 color difference $d_i$ is calculated from $\mathbf{I'}_c(x_i)$ and $\mathbf{\hat{I'}}_c(x_i)$. Then, the anomaly score is obtained simply by summing $d_i$ for all $i$. We briefly describe the concept of CIEDE2000 in the Appendix. For more details, see~\cite{Sharma_2005}.

\section*{\uppercase{Experiment}}
\subsection*{Dataset}
To evaluate the performance of the proposed method, we used a dataset that is publicly available through the PlantVillage project~\cite{Hughes_2016}. The dataset contains $54,306$ images of healthy and diseased plants covering $14$ crops: \textit{apples},  \textit{blueberries},  \textit{cherries},  \textit{corns},  \textit{grapes},  \textit{oranges},  \textit{peaches},  \textit{bell}  \textit{peppers},  \textit{potatoes},  \textit{raspberries},  \textit{soybeans},  \textit{squash},  \textit{strawberreis}, and \textit{tomatoes}. Each image exhibits three different versions: \textit{RGB color}, \textit{grayscale}, and \textit{segmented}.

Fig.~\ref{fig:Dataset} shows examples of segmented images in the PlantVillage dataset for (a) healthy and (b) diseased leaves. In the experiments, we used segmented potato plant images comprising 152 healthy leaves and 1,000 leaves with early blight, which is caused by the fungus Alternaria solani. Symptoms appear on older leaves as small brown spots. As the disease progresses, it spreads throughout the leaf surface and eventually makes it turn yellow and then wither.
%
\begin{figure}[htb]
\captionsetup{justification=raggedright}
\centering
\includegraphics[width=0.8\hsize]{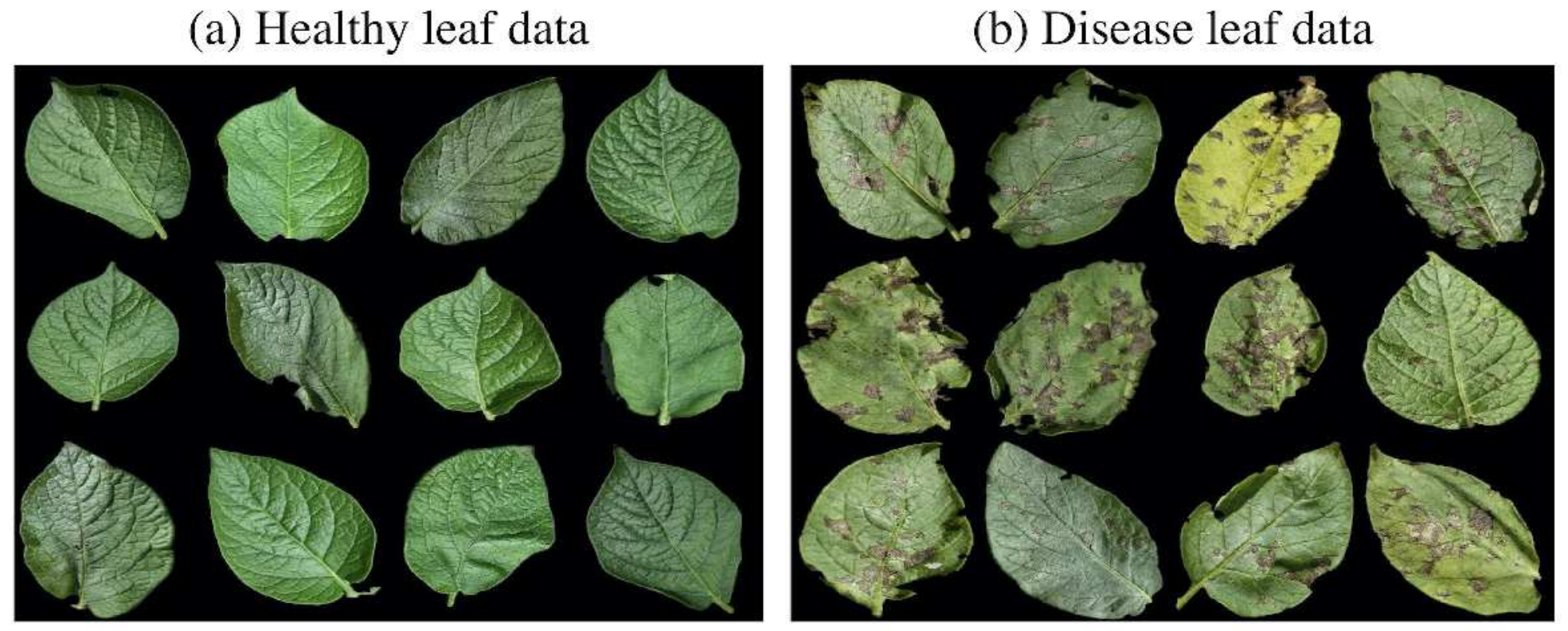}
\caption{Examples of potato plant segmented images in the PlantVillage dataset. (a) Healthy leaf Data, (b) Diseased leaf Data.}
\label{fig:Dataset}
\end{figure}

We divided these images into training and test sets for pix2pix, as given in Table~1.
Half of the healthy-leaf images were allocated to the training set, and half were allocated to the test set. We randomly chose 100 diseased-leaf images for the test set.
All images demonstrated a resolution of 256 $\times$ 256 pixels.

\setlength{\tabcolsep}{4pt}
\renewcommand{\arraystretch}{1.5}
\begin{table}
 \caption{Number of images used in the experiment.}
 \label{tab:dataset}
    \begin{center}
        \begin{tabular}{ccc}
            \hline
            Training data              & \multicolumn{1}{l}{Healthy leaf images} & 76  \\ \hline \hline
            \multirow{2}{*}{Test data} & Healthy leaf images  & 76  \\ \cline{2-3}
                                        & Diseased leaf images                     & 100 \\ \hline
        \end{tabular}
    \end{center}
\end{table}
\setlength{\tabcolsep}{1.4pt}


\subsection*{Experimental Setup} 
Here, we briefly describe the experimental setup for training the GANs. The optimization problem described in Eq.~\ref{eq:optimal} was solved with the Adam optimizer at a learning rate of $0.00015$. The momentums were set to $\beta_1 = 0.5$, and $\beta_2 = 0.999$. The discriminator has a PatchGAN architecture with a patch size of $64 \times 64$. Training was terminated after $150$ epochs. The hyperparameter for $L_{1}$ regularization was set to $\lambda = 10$.

For comparison, we use AnoGAN and the conventional unsupervised anomaly detection methods: Convolutional Autoencoders (CAEs) with L2 and SSIM ~\cite{bergmann2018improving}, One-Class SVM (OC-SVM) ~\cite{scholkopf2002} as baseline methods for evaluating the performance of our proposed method.

AnoGAN was also trained with the Adam optimizer at a learning rate was $0.0001$. Momentums were set to be the same as those for pix2pix. AnoGAN training was terminated after $2,000$ epochs. The latent variable dimension in AnoGAN was set to $30$. To calculate the anomaly scores, the weight coefficients for the residual loss and discrimination score were set to $0.9$ and $0.1$, respectively. Also, we calculated simple color histogram similarity to measure the color differences aiming at highlighting a characteristic of CIEDE2000.

We used Pytorch (version $1.3.0$) code for pix2pix which is available at https://github.com/phillipi/pix2pix. We modified the Keras (version $2.2.4$) code for AnoGAN which is available at https://github.com/tkwoo/anogan-keras. All computations were performed on a GeForce RTX 2028 Ti GPU based on a system running Python 3.6.9 and CUDA 10.0.130.

\subsection*{Color Differences at the Pixel Level} 
\label{sec:Visualization}
Fig.~\ref{fig:Colored_result} shows example results for the pixel-level visualization of test data. The top row (Healthy) shows the results for healthy potato leaves. The middle and lower rows (Disease (1) and Disease (2), respectively)) show the results for two diseased leaf examples. Fig.~3(a) shows the original color images, and Fig. 3(b) shows the grayscale conversion. Fig. 3(c) shows the color images reconstructed by pix2pix. Fig. 3(d) shows heat maps visualizing the CIEDE2000 color difference between the original and reconstructed color images. Warm colors indicate a large color difference.

As shown in Figs.~3(c), and (d), the healthy-leaf images were successfully reconstructed. However, the symptomatic brown spots and yellow discoloration in the diseased-leaf images were not reconstructed. These results aligned with our expectations ;because we trained pix2pix only with heathy-leaf images, the generator network could not reconstruct colors in symptomatic regions. Consequently, symptomatic regions exhibited large CIEDE2000 color differences. In contrast, healthy leaves exhibited no significant color differences.

For comparison, Fig.~3(e) shows the images reconstructed by AnoGAN. The generator network was trained with healthly-leaf images just like pix2pix. Fig.~3(f) presents heat maps visualizing the residuals for each pixel between the original and reconstructed color images using the same format as in Fig.~3(d).

The leaves were clearly reconstructed, incompletely, which strongly affected the pixel-level residuals. In particular, artificially highlighted regions can be observed around the edges of leaves in both the healthy and diseased cases. In contrast, the Disease (2) images indicate that most symptomatic regions, with yellow discoloration were not highlighted in the heat map. These results suggest the limitations of AnoGAN for visualizing symptomatic regions of plant leaves at the pixel level.

The incomplete reconstruction was most likely caused by a lack of training data. Accordingly, the reconstruction would be improved by adding healthy leaf images to the training data. However, this should cause the generator network, to fail to reconstruct diseased-leaf images, which may generate substantive artifacts. Thus, our proposed method provides a more efficient pixel-level visualization of anomalies in images in comparison with AnoGAN. In addition, our proposed method works well even though only 76 healthy-leaf images were used for training.

\begin{figure*}[!t]
\includegraphics[width=\hsize]{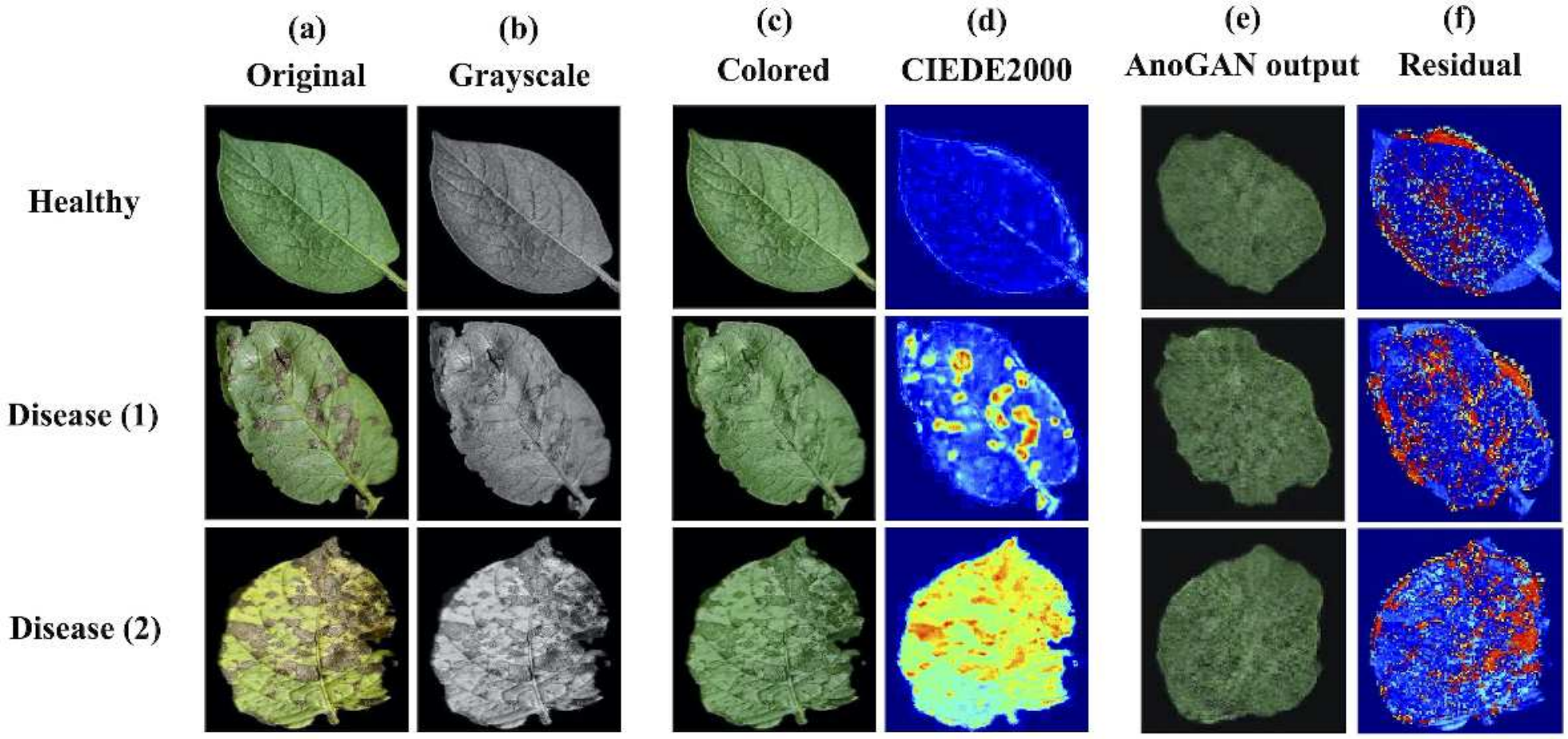}
\caption{Examples of pixel-level disease visualization of plants:
(a) original color images,
(b) grayscale conversion, and (c) color images reconstructed by pix2pix.
(d) CIEDE2000 color differences between the original color images and reconstructed color images.
(e) Images reconstructed by AnoGAN.
(f) Residuals between the original color images and images reconstructed by AnoGAN.}
\label{fig:Colored_result}
\end{figure*}

Fig.~\ref{fig:append} shows additional examples of pixel-level visualization of grape and strawberry leaf images with the proposed method. Similar to the previous results with potatoes, the symptomatic regions on these leaves were successfully highlighted. Thus, the proposed method works well at detecting symptomatic regions of various plants.

\begin{figure}[!t]
\centering
\includegraphics[width=\hsize]{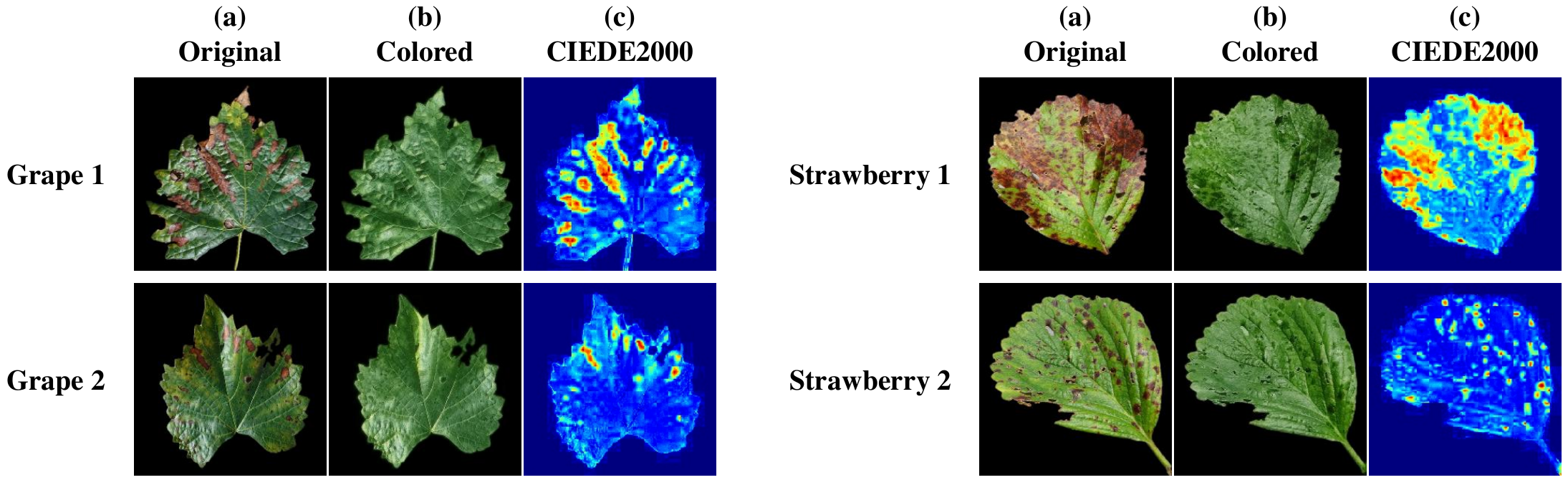}
\caption{Other examples of pixel-level disease visualization for grapes (first and second rows)
and strawberries (third and fourth rows):
(a) original color images, (b) reconstructed color images, and (c) CIEDE2000 color differences.}
\label{fig:append}
\end{figure}

\subsection*{Performance Evaluation of Image-level Disease Detection}
\label{sec:Evaluation}
Fig.~\ref{fig:Result} presents histograms for three anomaly scores:
(a) the CIEDE2000 anomaly score (i.e., proposed method),
(b) color histogram similarity, and (c) AnoGAN anomaly score.
The anomaly scores for healthy and disease-leaf images are indicated in red and blue, respectively.
The results indicate that the CIEDE2000 anomaly score is more intuitive for distinguishing healthy and diseased samples at the image level.

\begin{figure}[ht]
\centering
\includegraphics[width=\hsize]{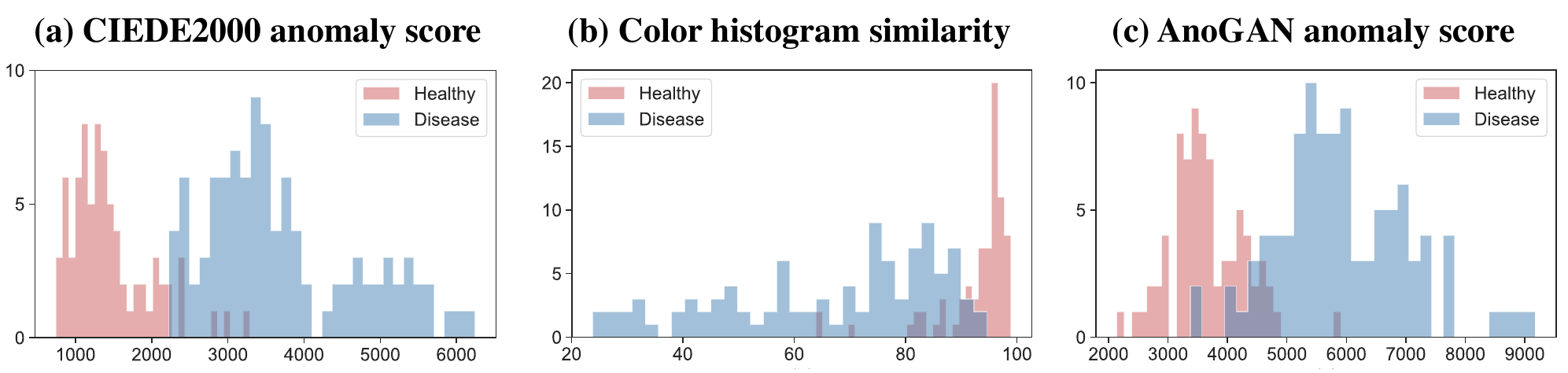}
\caption{Histograms of three anomaly scores.}
\label{fig:Result}
\end{figure}

To demonstrate the utility of the CIEDE2000 anomaly score in a more objective manner,
Fig.~\ref{fig:ROC}, shows receiver operating characteristic (ROC) curves for image-level disease detection with the three types of anomaly scores: CIEDE2000 (red), color histogram similarity (blue), and ANoGAN (green). The corresponding area under the ROC curves (AUC) is specified in parentheses in the figure legend. The shapes of the ROC curves indicate that the CIEDE2000 anomaly score demonstrates several useful properties: a high true positive rate and low false positive rate superior to those of the other two anomaly scores.

\begin{figure*}[ht]
\centering
\captionsetup{justification=centering}
\includegraphics[width=0.6\hsize]{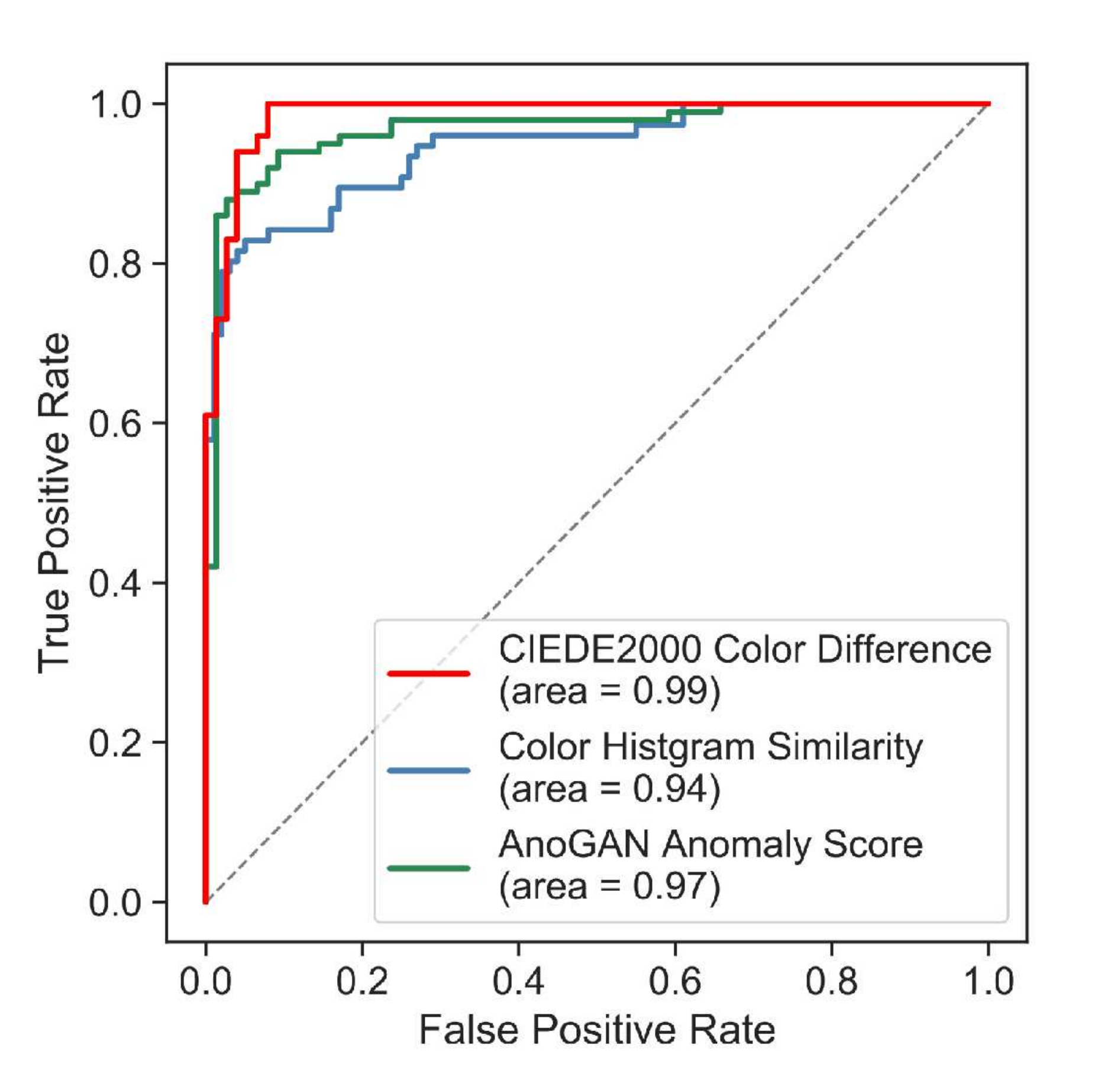}
\caption{ROC curves based on three anomaly scores.}
\label{fig:ROC}
\end{figure*}

Table~\ref{tab:result} presents the statistical performance of the proposed method at image-level disease detection in terms of the precision, recall, and $F_{1}$-score. These indices were calculated the top-100 test images sorted in decreasing order with respect to anomaly score. The best performance for each index is bolded. The CIEDE2000 anomaly score demonstrated a superior performance compared to the two baseline anomaly scores for all indices.

\setlength{\tabcolsep}{4pt} 
\begin{table}
    \begin{center}
    \caption{Comparison of image-level disease detection performances with the PlantVillage dataset.}
    \label{tab:result}
    \begin{tabular}{lllll}
    \hline\noalign{\smallskip}
    Method & ~ Precision ~ & ~ Recall ~ & ~ $F_{1}$-Score ~ & ~ AUC~  \\
    \noalign{\smallskip}
    \hline
    \noalign{\smallskip}

    CIEDE2000 Anomaly Score                 & ~ \textbf{0.94} ~ & ~ \textbf{0.94} ~ & ~ \textbf{0.94} ~ & ~ \textbf{0.99} ~  \\ 
    color Histogram Similarity & ~ 0.86 ~          & ~ 0.86 ~          & ~ 0.86 ~          & ~ 0.94 ~           \\ 
    AnoGAN Anomaly Score                  & ~ 0.92 ~          & ~ 0.92 ~          & ~ 0.92 ~          & ~ 0.97 ~           \\ 
    AutoEncoder L2 Loss                  & ~ 0.67 ~          & ~ 0.68 ~          & ~ 0.69 ~          & ~ 0.67 ~           \\ 
    AutoEncoder SSIM Loss                  & ~ 0.89 ~          & ~ 0.90 ~          & ~ 0.90 ~          & ~ 0.94 ~           \\ 
    OneClass SVM                  & ~ 0.91 ~          & ~ 0.92 ~          & ~ 0.92 ~          & ~ 0.95 ~           \\ 
    \hline
    \end{tabular}
    \end{center}
    \end{table}
    \setlength{\tabcolsep}{1.4pt}

\subsection*{Computational Efficiency for the Anomaly Score}
During detection, AnoGAN needs to determine the latent space location for a given query image based on iterative backpropagation, which leads to the anomaly score~\cite{Schlegl2017}\cite{Schlegl_2019}. This process reduces the computational efficiency of AnoGAN. In contrast, our proposed method does not require an iterative procedure during detection. Table~\ref{tab:computation} presents the mean computation times in milliseconds with standard deviations for calculating the CIEDE2000 anomaly score and AnoGAN anomaly score of a given query image. The results suggest that our approach offers superior computational efficiency compared to AnoGAN, which is clearly an important consideration for the practical application of real-time disease detection in plantations.

\setlength{\tabcolsep}{4pt}
\renewcommand{\arraystretch}{1.5}
\begin{table}
    \caption{Computation times for anomaly scores.}
    \begin{center}
        \begin{tabular}{ccc}
            \hline
                 & \multicolumn{1}{l}{\textit{Healthy}} & \multicolumn{1}{l}{\textit{Diseased}} \\ \hline
            Our method & $61.24 \pm 17.21$~ms                                 & $62.09 \pm 18.33 $~ms                                 \\
            AnoGAN     & $3884 \pm 1138 $~ms                                  & $4574 \pm  1538 $~ms                                  \\ \hline
        \end{tabular}
    \end{center}
     \label{tab:computation}
\end{table}
\setlength{\tabcolsep}{1.4pt}

\section*{\uppercase{Conclusions}}
We proposed a novel method for detecting plant diseases from images that relies on color reconstructability. Similar to AnoGAN, the proposed method detects anomalies in test data based on unsupervised training of a generator and discriminator by normal data. Unlike AnoGAN, however, the proposed method predominantly focuses on color anomalies in images.

We compared the performance of the proposed method with baseline methods including AnoGAN in terms of accuracy, interpretability, and computational efficiency. Experiments with the PlantVillage dataset showed that the proposed method performed better than AnoGAN at image-level anomaly detection. Because the CIEDE2000 anomaly score is simple and aligns with human visual inspection, it can be intuitively visualized as a heat map at the pixel level. In representative examples, symptomatic regions on leaves such as brown spots and yellow discoloration were efficiently highlighted. No serious artifacts were observed in either healthy- or diseased-leaf images, in contrast to the residual maps based on AnoGAN.

Because the proposed method does not require any iterative computation for calculating anomaly scores, the mean computation time is significantly less than that of AnoGAN. The computational efficiency means that the proposed method could be practical for application to real-time image-based plant disease detection. Future studies are warranted to explore the practicality of automatic diagnosis systems for detecting plant diseases on a global scale based on the idea of color reconstructability.

\section*{\uppercase{Acknowledgments}}
This work was supported by JST,  PRESTO Grant Number JPMJPR1875, Japan. The authors would like to thank Enago (www.enago.jp) for the English language review.

\clearpage

\section*{\uppercase{Appendix}}
\subsection*{CIEDE2000 Color Difference}
The CIEDE2000 color difference is calculated by using the color space $L^*a^*b^*$, which is suitable for
expressing colors based on human perception. The color difference is based on three parameters:
the lightness difference ($\Delta L^{\prime}$), chroma difference ($\Delta C^{\prime}$) and hue difference ($\Delta H^{\prime}$). These are weighted by the functions ($S_L$, $S_C$, $S_H$), parametric weighting factors ($k_L$, $k_C$, $k_H$), and rotation term ($R_T$). All parametric weighting factors were set to $k_L = k_C = k_H = 1$.

The CIEDE2000 color difference between two points in the $L^*a^*b^*$ color space $(L^*_1,a^*_1,b^*_1)$ and $(L^*_2,a^*_2,b^*_2)$ is calculated
as follows:
\begin{align}
    &\Delta E_{00}(L^*_1,a^*_1,b^*_1,L^*_2,a^*_2,b^*_2)= \nonumber \\ \nonumber \\
    &\sqrt{\left(\tfrac{\Delta L^{\prime}}{k_{L} S_{L}}\right)^{2}+\left(\tfrac{\Delta C^{\prime}}{k_{C} S_{C}}\right)^{2}+\left(\tfrac{\Delta H^{\prime}}{k_{H} S_{H}}\right)^{2}+\left(\mathrm{R}_{\mathrm{T}}\left(\tfrac{\Delta C^{\prime}}{k_{C} S_{C}}\right)\left(\tfrac{\Delta H^{\prime}}{k_{H} S_{H}}\right)\right)}
    \label{eq:CIEDE2000}
\end{align}
The Python code used for implementing this formula is available at https://github.com/scikit-image. Further details and a derivation of the CIEDE2000 color difference equation are provided by~\cite{Sharma_2005}.

\subsection*{Application for MVTec Anomaly Detection Dataset}
We validated the effectiveness of our anomaly detection method for industrial problems by using the MVTec Anomaly Detection Dataset (MVTec AD)~\cite{Bergmann_2019}. This is a comprehensive dataset for benchmarking anomaly detection methods with a focus on industrial applications. It contains 5,354 high resolution color images of different object and texture categories with annotated pixel-level ground-truth regions for all anomalies. Fig.~\ref{fig:MVTec} shows examples of pixel-level anomaly visualizations with the proposed method. The top row shows the input images. We selected five categories from MVTec AD whose anomalies affect the color. The proposed method effectively highlighted anomalies on industrial products.

\begin{figure}[hbt]
\centering
\captionsetup{justification=centering}
\includegraphics[width=0.9\hsize]{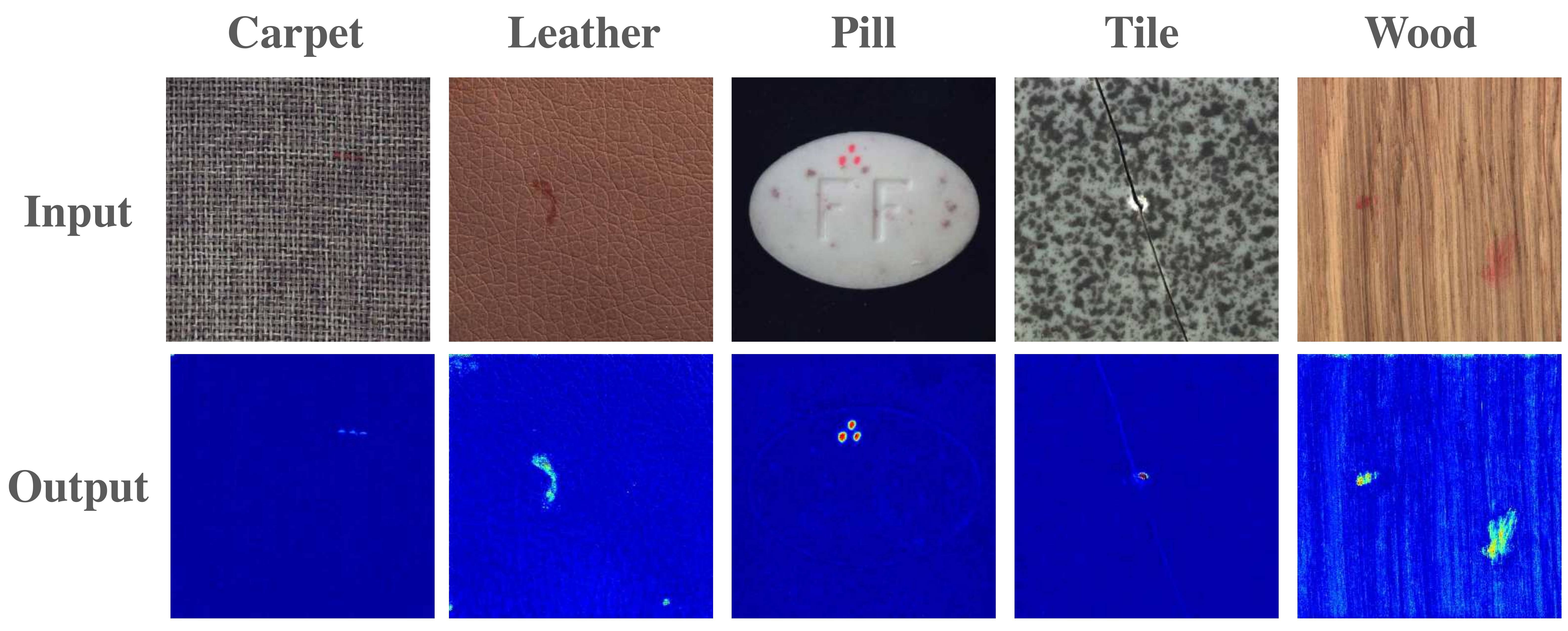}
\vspace{0.05em}
\caption{Examples of pixel-level anomaly visualization with MVTec AD.}
\label{fig:MVTec}
\end{figure}

\end{document}